\documentclass[10pt, a4paper]{article}
\usepackage{lrec-coling2024} 
\usepackage{graphicx}
\usepackage{caption}
\usepackage{subcaption}
\usepackage{comment}
\title{Word Ladders: A Mobile Application for Semantic Data Collection}
\name{Marianna Marcella Bolognesi, Claudia Collacciani, Andrea Ferrari, \\ {\bf \large Francesca Genovese,
Tommaso Lamarra, Adele Loia,} \\
{\bf \large Giulia Rambelli, Andrea Amelio Ravelli, Caterina Villani}}
\address{University of Bologna, Italy \\
          \\
         Corresponding: m.bolognesi@unibo.it\\}
\abstract{
Word Ladders is a free mobile application for Android and iOS, developed for collecting linguistic data, specifically lists of words related to each other through semantic relations of \textit{categorical inclusion}, within the Abstraction project (ERC-2021-STG-101039777). We hereby provide an overview of Word Ladders, explaining its game logic, motivation and expected results and applications to nlp tasks as well as to the investigation of cognitive scientific open questions.
\\ \newline
}
\begin{document}
\maketitleabstract
\vspace{0.5\baselineskip}
\section{Introduction}
The adoption of gamified methodologies for collecting human data is considered a promising path for developing linguistic resources, such as the construction of thesauri and ontologies. These resources, in turn, can be used in NLP tasks \cite{jurgens2014s, madge2017testing, bonetti2021measuring}. 

The present paper briefly illustrates the rationale, functioning, technical structure of the mobile application Word Ladders, freely available on Apple and Android devices
\footnote{Android download: \href{https://t.co/6ZXADJhRuq}{http://shorturl.at/kqAOS}

Apple download: \href{https://t.co/QMHmMgKL1x}{http://shorturl.at/buIS2}}, 
developed by the Abstraction research group as part of a 5-year project funded by the European Research Concil (ERC-2021-STG-101039777). 

\begin{figure}[t!]
    \centering
    \includegraphics[width=.5\textwidth]{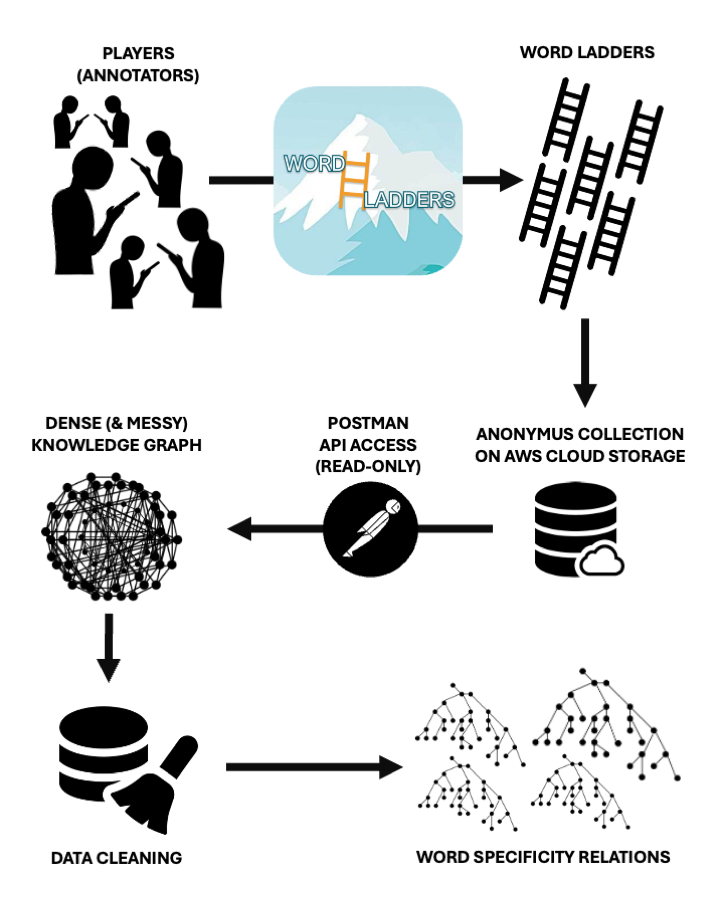}
    \caption{Schematic representation of the workflow of data collection and processing. (a) Data are collected through Word Ladders, anonymized and stored into a AWS server. (b) Stored data are accessed using Postman API and converted to generate a graph. (c) The resulting graph is post-processed to detect typos and remove noisy ladders. (d) The final graph is used to i.) understand the semantic organization of users (of different sociodemographic backgrounds) and ii.) compute the Specificity rating for the given words.}
    \label{fig:ANON GAME_workflow}
\end{figure}

Word Ladders has been developed with a scientific and educational objective in mind. On the research side, it is designed to collect linguistic data, specifically lists of words related to each other through semantic relations of \textbf{categorical inclusion }or meaning extension, also referred to as \textit{hypernymy}/\textit{hyponymy} in linguistics, or \textit{IS-A relation}. For example, given the word \textit{apple} add increasingly more generic words such as \textit{fruit} and \textit{food}, and more specific ones, such as \textit{Fuji}. The pool of words used in Word Ladders is a subset of the 40.000 words presented in the English dataset of concreteness ratings collected by \citet{brysbaert2014concreteness} and, for which several other psycholinguistic metrics have been collected in rating tasks through the past decade. In particular, the words that compose the Brysbaert database have been filtered and also manually checked, to include only words that do not present potential violent connotations, swearwords and other words that may be sensitive, based on the guides offered by Ipsos MORI in the “Public attitudes towards offensive language on  TV and Radio: Quick Reference Guide” commissioned by Ofcom. The mobile application therefore takes into account sensitivities of players with different ages, including minors.
The data collected through this application is intended for constructing an ontology containing hierarchical semantic relations of category inclusion as well as sociodemographic data about the users that produced such associations (please see below for more details about the sociodemographic data collected by the app, in anonymized form). From this resource, it will be possible to extract metrics of word Specificity for words in English and Italian. These measures, in turn, will be used to predict word processing and comprehension together with other psycholinguistic variables, and to explore how lexical precision affects readability for different types of texts, written for different types of audience. 

The scientific objectives tackled by the academic programme that supported the development of Word Ladders are described in detail on the Abstraction website\footnote{\url{https://www.abstractionproject.eu/}}. 

On the user side, the game aids speakers in doing "linguistic stretching", prompting them to recall highly specific and highly generic words from memory, and connecting them to basic level vocabulary, thanks to the feedback that the application provides to users after each played match. The process of "linguistic stretching" helps users keep their active lexicon trained. The application was published on mobile stores in early September 2023 and can be freely downloaded.

\section{Related Works
}
Gamification implies the integration of game design elements into non-game contexts to stimulate users' participation \cite{terril2008gamification}. This principle has proven effective in serious games, which are interactive computer-based tools designed to achieve educational outcomes such as learning new information or developing particular skills \cite{ritterfeld2009serious}. In Natural Language Processing (NLP), where user engagement and data collection are paramount, the application of gamification techniques is typically implemented through GWAPs (games with a purpose), which are a type of game in which the player’s actions contribute to scientific or anyway real-world purposes, such as semantic annotations, image labeling and so forth (e.g., \citealp{von2008designing}). This approach holds significant potential to scale up data collection that is otherwise boring for participants, in addition to being time and energy-consuming for researchers \cite{mcpherson2007gs, deright2015just}. 

In language sciences, recent research indicates promising prospects for gamification in enhancing engagement, proficiency, and interaction in language learning and communication processes \cite{smith2022games, shortt2023gamification}, second-language acquisition, grammar proficiency, and advanced linguistic issues \cite{benini2021critical, dehghanzadeh2021using, shortt2023gamification, bonetti2021measuring}. However, the effectiveness of gamified techniques is not consistently proven \cite{lumsden2016gamification}. Studies show advantages of gamified approaches with participants with certain clinical conditions, such as ADHD (Attention Deficit Hyperactivity Disorder; \citealp{lumsden2016gamification, smiderle2020impact}), or for particular tasks such as training healthy behaviors and environmental awareness \cite{qian2016game, hammady2022serious}. Yet, in other cases, the addition of game elements may not significantly enhance performance or engagement and could introduce uncontrolled variables \cite{gundry2019validity}. Therefore, the decision to gamify or not gamify depends on the specific user population, task nature, and the potential for enhanced engagement and performance.
In the particular case of linguistic data for NLP tasks, a gamified approach designed to collect linguistically relevant data can hardly be realized within the framework of a highly recreational game, because the information requested from players involves the commitment to provide a substantial cognitive effort, which typically discourages users attracted by easy and fast entertainments and rewards. We address this challenge arguing that the benefits gained by users may target their cognitive training, the enhancement of their linguistic proficiency, and similar educational outcomes. We, therefore, introduce Word Ladders, a mobile application characterized by a dual objective: one oriented towards fulfilling scientific objectives for researchers and another focused on users' linguistic training, achieved alongside the app's entertaining appeal. In the present contribution Word Ladders is illustrated, and the challenges involved in relation with the two objectives are discussed.

\section{WORD LADDERS}

\subsection{Game Logic}

The game revolves around eliciting a particular type of word association, namely the semantic relation IS-A. Users are encouraged to create word ladders by adding steps above and below an initial prompt. The game can be played in English or Italian.

The game's rules, presented to players, are as follows:
\begin{enumerate}

\item The primary aim is to construct a ladder by adding words on each step, but not all words yield points, so make thoughtful choices.

\item Commence by adding progressively more general words above the provided prompt. For instance, if the starting word is \textit{APPLE}, you can add \textit{fruit} above it, then \textit{food} above \textit{fruit}, and finally \textit{object} above \textit{food}. The key is to think, "The word I just added describes a type of..."

\item Ensure that each added word is more general than the previous one. In the example, APPLE is a type of FRUIT, which is a type of FOOD, which is a type of OBJECT.

\item Conversely, you can be more specific and precise on the steps below the initial word. For instance, you can add \textit{Granny Smith}, a specific type of apple, below APPLE.

\item Accumulate more points by creating longer ladders, so challenge yourself and your friends to build the lengthiest ladder possible.

\item To aid word selection, consider the category or type that the placed word belongs to.

\item If unsure about how to play, a graphic tutorial is available to guide players through the gameplay and clarify the rules.
\end{enumerate}

The Word Ladders platform offers three distinct game modes: individual, challenge, and group. The objective remains the same in all three modes: building the longest ladder of valid words, starting from a given prompt. The time allocated to complete a match (namely: a ladder) is uniformly set at 120 seconds across all modes. This time allocated has been established based on experimental sessions followed by debriefing sessions conducted at the hosting institution, and reported in \citep{genovese_gamification}.

At the end of the time allocated for the match, the proposed ladder is automatically rated on a scale of 1 to 5 stars. 
The goodness of ladders is assessed against an external knowledge base that features the semantic relation IS-A, namely MultiWordNet 1.5.0 \cite{MultiWordNet_2002}, a multilingual lexical database\footnote{Licensed under a Creative Commons Attribution 3.0 Unported License} aligned with Princeton WordNet 1.6, hosted by FBK Foundation in Trento (Italy). Of the initial pool of words (Brysbaert's dataset of concreteness ratings) filtered for sentitive words, only words listed in MultiWordNet are retained, with the exact number of words in English and Italian, respectively.  To evaluate ladder quality and assign user ratings and scores, two parameters are considered: the number of entered words and the validity of those words (namely: whether they are in MultiWordNet 1.5.0). Additionally, a graph of the words produced by users is constructed in the backend. Therefore, if users frequently produce a word in response to a starting prompt (e.g., \textit{Fuji}' produced below \textit{apple}), it becomes a valid entry even if it is not present in MultiWordNet (after a production frequency of N=50).

Word Ladders functions as a gamified method for collecting linguistic data that can be used for NLP tasks, as described below. Additionally, it qualifies as a serious game by not only incorporating gamification elements into scientific tasks but also aiming to enhance and broaden players' vocabulary. In a recent paper, \citep{genovese_gamification} provide a case study in which linguistic associations of the type IS-A are collected in different experimental conditions: 1. using a beta version of Word Ladders in Individual mode; 2. using a beta version of Word Ladders in Challenge (1 vs. 1) Mode, and 3. using a classic online survey form (control group). Results show that the quality of linguistic data acquired through Word Ladders is comparable with that obtained through traditional survey-based methods. Moreover, the study also shows that users' motivation to complete the task and possibly repeat it in the future is significantly higher when the task is performed through Word Ladders, compared to the control group that used a standard survey. This was particularly true for players who performed the task using the Challenge mode of the app, which shows that the interactive aspect and the 1-to-1 competition increased users' engagement. This suggests that Word Ladders is a scalable approach to collect large quantities of linguistic data.

\subsection{Motivations}

\subsubsection{Relevance for Language Sciences and NLP}

The theoretical problem that led to the implementation of Word Ladders is hereby explained. When working on conceptual abstraction from cognitive and computational perspectives, it is often implied that two different variables are correlated, namely: concreteness/abstractness (\textit{banana} vs \textit{belief}) and specificity/genericity (\textit{banana} vs \textit{food}). The assumption goes in the direction that abstract (intangible) concepts are also more generic than concrete ones because these two variables, which for simplicity are called \textbf{Concreteness} and \textbf{Specificity}, are thought to be highly correlated. However, recent studies \cite{bolognesi2020abstraction,bolognesi2023specificity} demonstrate this is not the case: the two variables are theoretically distinct and function as separate significant predictors of linguistic performance in lexical and semantic decision tasks. While the cognitive scientific community and the NLP community are now directing their attention to these two linguistic variables, when it comes to Specificity, the resources available to operationalize and measure it are very limited, often metrics based on WordNet.

WordNet\cite{miller1995wordnet} is a lexical resource that conveniently encodes the semantic relation IS-A (e.g., \textit{cat IS A type of mammal}), allowing for the extraction of specificity metrics (as in \citealp{bolognesi2020abstraction}). However, the vocabulary known by WordNet is much larger than the lexical knowledge of any speaker. For instance, under the node INSECT, WordNet allows up to six subordinate levels before reaching a leaf node (e.g., insect, beetle, lamellicorn beetle, scarabeid, dung beetle, scarabaeus sacer). Because of the richness of WordNet's encyclopedic knowledge, specificity scores extracted from this taxonomy do not reflect speakers' lexical knowledge and allow for only partial investigations into the cognitive mechanisms of abstraction and generalizations in language. 

For this reason, \citet{bolognesi2023specificity} and \citet{specificity_en} collected specificity scores from speakers in an experimental setting, using a Best-Worst Scaling paradigm \cite{louviere2015best} through which participants were asked to indicate the most and least generic word among tuples of four. For instance, given four nouns like

\begin{enumerate}
    \item[] \textit{reluctant, surprised, shy, emotional}
\end{enumerate}
the participant should select \textit{reluctant} as the more specific and \textit{emotional} as the most generic term. 
The Best-Worst Scaling provides a helpful method for constructing resources to measure Specificity. However, this paradigm is time and energy-consuming, and does not allow exploration of speakers' variability in the production of categorizations, because the words are arbitrarily selected by the experimenters.

Conversely, Word Ladders is a generative task inviting users to construct generalizations through a gamified approach. The application collects sociodemographic data about users (which are not connected to personal information) so that it is then possible to analyze the type of word ladders produced in relation to age, profession, level of education, first language, and reading habits. For each of these variables, it is possible to develop hypotheses about their impact on the length and the accuracy of the ladders produced by players. For instance, it can be argued that age, level of education, and reading habits have positive effects on the length and accuracy of word ladders produced by players. Crucially, the sociodemographic information collected upon registration in the app is also used within the game to enable players to measure their performance with their peers, filtering the leaderboard based on similar types of players. 

The word ladders collected through the application will be used to construct the following two main resources:

\begin{itemize}
    \item a dataset of specificity scores, built by extracting specificity directly from word ladders, measuring the position of a word within a ladder divided by the ladder length;
    \item a taxonomy of hierarchically organized categories, in which the primary semantic relation encoded is the semantic relation IS-A of category inclusion (but not limited to this, given that players tend to insert other hierarchical relations, such as part-whole).
\end{itemize}

The resources listed above enable a range of downstream tasks. Primarily, with specificity scores available for both Italian and English words, we can delve into the role of Specificity in language processing. This includes the investigation of Specificity in language-related tasks such as lexical decisions, recall, semantic decisions, production, and memorization, in which linguistic behavior is explained by the levels of word Specificity.
Additionally, these scores are required to evaluate various text-related tasks, such as assessing text readability and comprehensibility based on lexical features, which may include specificity. Furthermore, integrating specificity scores into assisted-writing software allows for suggestions of word replacements aimed at enhancing text clarity. 

Finally, due to the open, generative nature of this linguistic game, which involves semantic categorizations on various levels of inclusiveness, it would be possible to observe the emergence of language stereotypes. In fact, during a public engagement event where the present application was disseminated, gender stereotypes emerged spontaneously from players asked to construct word ladders starting from the words \textit{doctor} (some indicating \textit{man} as an IS-A relationship) and \textit{caregiver} (some indicating \textit{woman} as an IS-A relationship). Given that Word Ladders collects some players' sociodemographic information, it will be possible to associate the emergence (and decline) of stereotypes with different types of players.

\subsubsection{Educational Applications}

Word Ladders has also been implemented as an educational tool in public schools, aimed at supporting students' language training and vocabulary-related activities. Beyond its widespread adoption in Italian public schools, where it has been introduced and utilized in 84 classes spanning from 5th grade (primary school) to high school across three regions, the app is currently used in a longitudinal case study at Imola's IC7 primary and middle schools, Italy. Specifically, Word Ladders is integrated into three phases of classroom interventions—beginning, middle, and end of the school year—with two groups of parallel classes employing distinct teaching methodologies for vocabulary development. The length and accuracy of the word ladders produced by pupils will be compared between the two groups, each employing a different methodology for lexical training: one group using a traditional method and the other an experimental approach that may give pupils an advantage in developing fine-grained lexical knowledge.

\section{App Documentation and Data Collection} 

\subsection{Service Implementation}

Word Ladders is based on a React Native\footnote{\url{https://reactnative.dev}} front-end, a NodeJS\footnote{\url{https://nodejs.org}} back-end, and a MongoDB\footnote{\url{https://www.mongodb.com}} database. The code then underwent a refactoring phase with an external gaming studio [Anon], which also finalized the publication of Word Ladder on the mobile stores. The application's back-end is hosted on AWS cloud server and is accessed through APIs, some of them used by the application's front-end, and others developed to grant data access to researchers (using Postman\footnote{\url{https://www.postman.com}}). 

For the researchers involved in the programme, it is possible to download the collected data through the APIs.  The access is limited to an individual account, protected by a strong password (alphanumeric). The filtering API allows the selection of the desired MongoDB collection, the specification of a MongoDB query, and the output file format (CSV or JSON).
The raw API allows the download of the computational tree used by the system to track and validate the users' ladders.

Data collected through the application are stored through the back-end architecture into a database, which is a cloud server hosted on Amazon AWS, on which the Abstraction research group have a dedicated account protected by strong password. The Computing machine within the AWS cloud server is private and dedicated to the project, on a server hosted within the European territory. Data collected from users and stored on AWS include the following information provided by the registered users: age, level of education, profession, mother tongue. These are associated with the username, which is an invented nickname. The application does not collect personal information such as email, phone number, IP, or geolocation. Data from crash/non-fatal error management platforms are recorded, such as device model and operating system on which the problem occurred. However, this data is not associated with the user. This information is used by the development teams to correct any product problems but not to monitor the user's activities. The library that does this is Google's Firebase Crashlytics. In summary: some device data is collected but is not associated with users, the data is collected only in case of problems and is not for monitoring purposes.

\paragraph{Game levels } 
The pool of words from which Word Ladders randomly draws an item for each match is divided into 50 levels of increasing difficulty, which are based on a series of nested rankings that take into account word frequency (the more frequent the easier), word concreteness (the more concrete the easier), word familiarity (the more familiar the easier) and part of speech (nouns before verbs, before adjectives). Players are exposed to 10 words randomly selected for each level, before they move automatically to the next level, provided that they reached a threshold of ladder accuracy. The levels are created based on word concreteness, frequency, and familiarity. Therefore, users will play on a word like \textit{banana} or \textit{wolf}, before they get to construct a ladder on words like \textit{entropy} or \textit{insinuation}. As previously mentioned, words have also been automatically filtered and subsequently manually checked for appropriateness. All words referring to taboo domains, violent domains or sensitive domains have been removed, since the application is used by minors as well as by adults. Guidelines provided by a dedicated Ethics Consultant and by the Ethics' committee of the hosting institution were followed (Protocol of Approval n. 0121435, May 20th 2022).

\paragraph{Data structure} To process the data of a ladder game and evaluate the validity and the score of the ladder, the designed structure is a graph that always has a center starting point which is the starting word of the ladder. The graph can be described as the union of the different graphs that start from the node that represents the starting word of the ladder, one for representing the hypernyms and the other one to represent the hyponyms side of the ladder. Every node can have multiple connections, the only connection that is not allowed is between a node of the hypernyms graph and the hyponyms one. This structure is pre-generated starting from the knowledge base of MultiWordNet and will be populated with the data of every play. Every arc that connects two nodes contains information about how any time other users generated that arc or if was present inside MultiWordNet. This graph structure is compressed into a list of nodes and two lists of arcs, one for hypernyms and the other for hyponyms.

\paragraph{Ladder evaluation} 
To evaluate the ladder built by the user, both words attested in MultiWordNet (i.e., validated words) and novel terms (i.e., not validated words) are considered. 
The validated ladder quality is proportional to the amount of play on the same ladder. The number of plays will be used to balance the two ladder lengths, validated and not validated (the original ladder sent by the user).
Defining: \textit{np} as the number of plays of the ladder for a word \textit{w}, \textit{r} is a constant value comprised between 0.2 and 0.8 referring to the ratio that will balance the two ladder lengths, and \textit{g} as the number of plays needed to obtain a good evaluation of the ladder, then 
\begin{equation}
    npl=min(max(\frac{np}{g}, 0.2), 0.8)
\end{equation}
and the score \textit{s} of the user ladder is computed as 
\begin{equation}
    s=100*npl*\frac{min(ulv,m)}{m}+(1-npl)*\frac{min(ul,m)}{m}
\end{equation} 
where \textit{m} is the greatest ladder length ever achieved, \textit{ul} is the length of the original user ladder, and \textit{ulv} is the length of the validated user ladder. After that, the score can be increased up to 10 points as the time spent by the user in the game decreases, the upper bound of the score is always 100.
To calculate \textit{ulv}, all the words and their position inside the user ladder are considered with the already known possible ladders, and those among them that respect the minimum threshold will be considered valid.

\subsection{Preliminary Data Analyses}


\begin{table}[htbp]
\begin{center}
\begin{tabular}{rl}
\hline
\multicolumn{2}{c}{\textbf{USERS}}        \\ \hline
Number of users          & 3196           \\
Age mean (std)           & 33.51 (19.77)  \\ \hline
\multicolumn{2}{c}{\textbf{PLAYED GAMES}} \\ \hline
Single player            & 28223          \\
Challenge                & 5142           \\
Team                     & 531            \\
Average games per user   & 10.62          \\ \hline
Italian                  & 31825          \\
English                  & 2115           \\ \hline
\textbf{TOTAL}           & 33896         
\end{tabular}
\end{center}
\caption{Number of users and played games in Word Ladders; data dump of February 13th, 2024.}\label{table:played}
\end{table}

\begin{table*}[htbp]
\centering
\resizebox{\textwidth}{!}{%
\begin{tabular}{rl}
\hline
\multicolumn{2}{c}{\textbf{Good ladders}}                              \\ \hline
\textbf{IT} & essere vivente, animale, vertebrato, mammifero, canide, \textbf{VOLPE}, volpe artica     \\
\textbf{EN} & living being, animal, mammal, canine, \textbf{FOX}, grey fox      \\ \hline
\multicolumn{2}{c}{\textbf{Bad ladders}}                               \\ \hline
\textbf{IT} & formaggio, latte, mammifero, felino, gatto, \textbf{VOLPE}, furbo \\
\textbf{EN} & world, solar system, planet, earth, race, species,  apes,  monkey,  animal, \textbf{FOX}
\end{tabular}%
}
\caption{Examples of data collected with Word Ladders. On the top, examples of good ladders produced in Italian and English with the starting word \textit{fox} (\textit{volpe}, in Italian); on the bottom, two ladders with the same starting words that do not express a specificity relation. }
\label{tab:word_ladders_examples}
\end{table*}

\begin{figure*}
\centering
\begin{subfigure}{.5\textwidth}
  \centering
  \includegraphics[width=\textwidth]{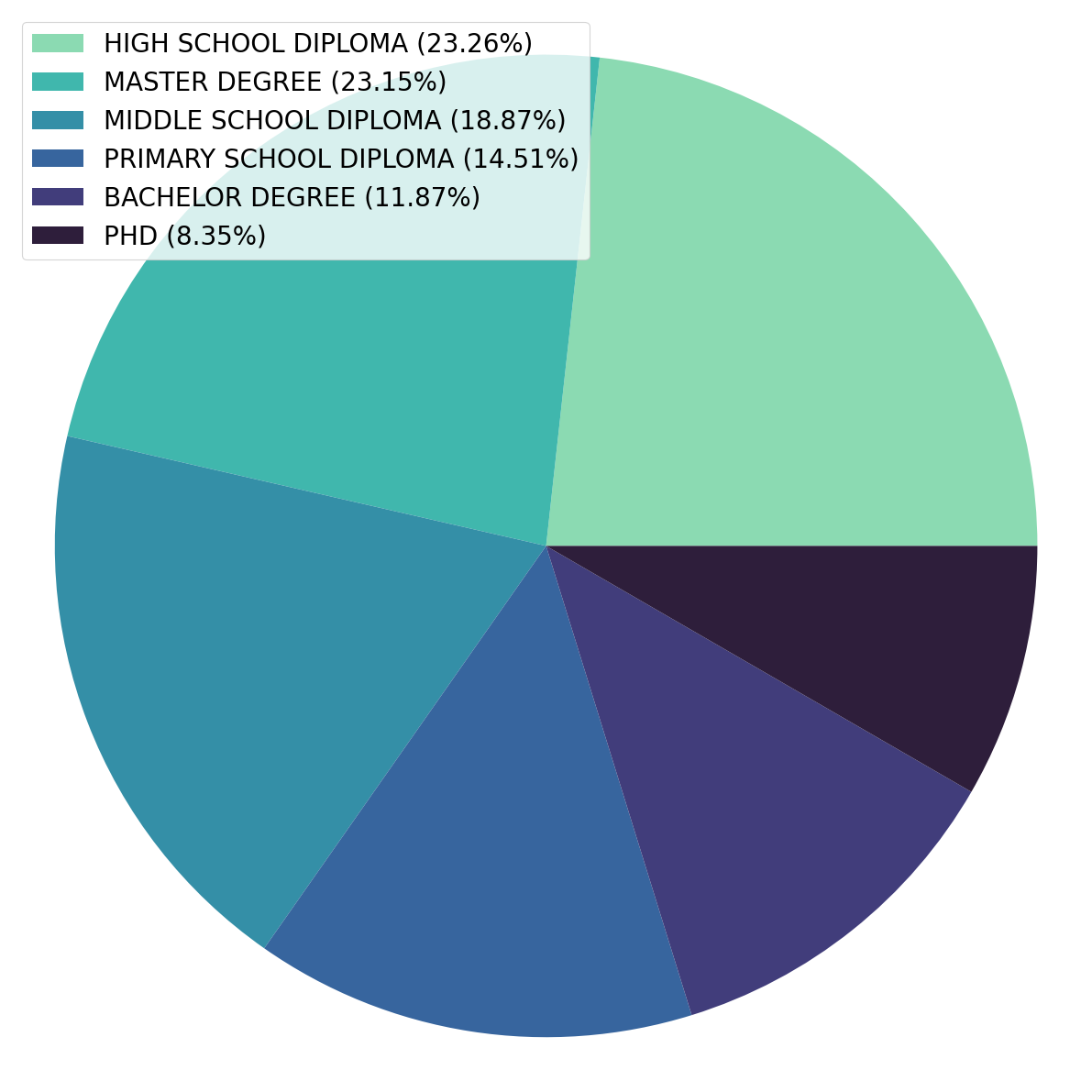}
  \caption{Highest education level.}
  \label{fig:sub_edu}
\end{subfigure}%
\begin{subfigure}{.5\textwidth}
  \centering
  \includegraphics[width=\textwidth]{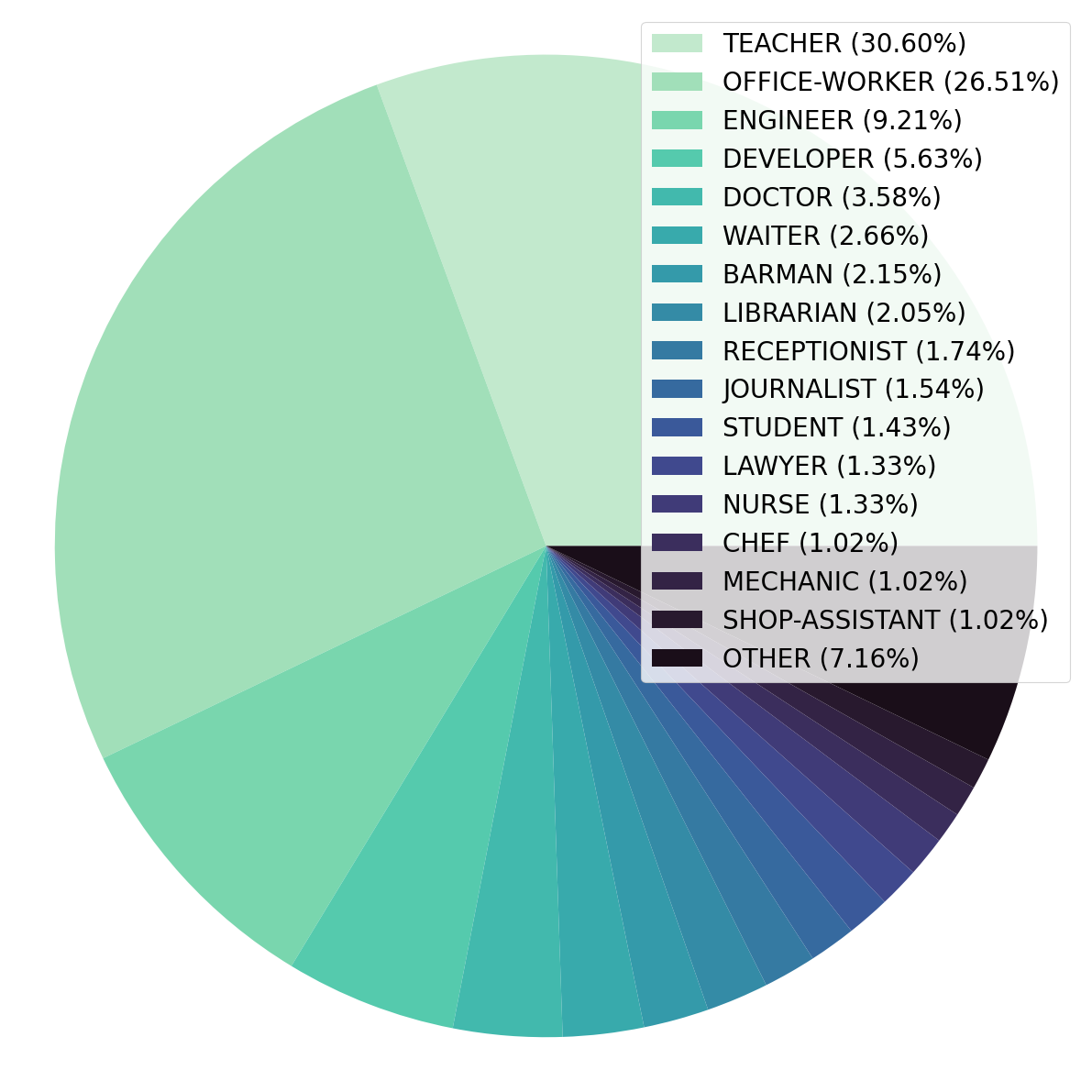}
  \caption{Profession.}
  \label{fig:sub_prof}
\end{subfigure}
\caption{Education (a) and profession (b) information about Word Ladders users.}
\label{fig:pies}
\end{figure*}

\begin{figure*}
\centering
\begin{subfigure}{.5\textwidth}
  \centering
  \includegraphics[width=1.003\textwidth]{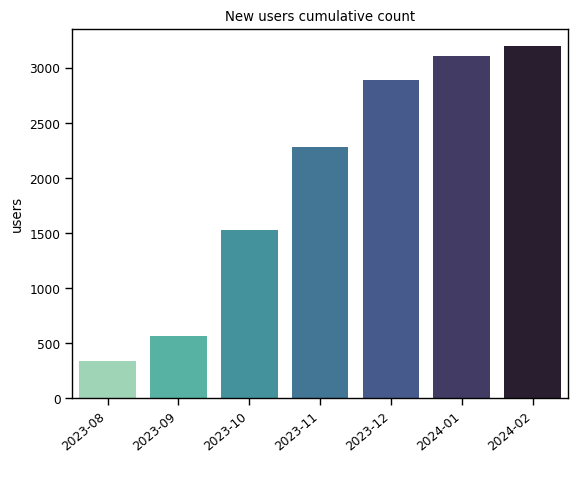}
  \caption{New users count per month.}
  \label{fig:sub3}
\end{subfigure}%
\begin{subfigure}{.5\textwidth}
  \centering
  \includegraphics[width=\textwidth]{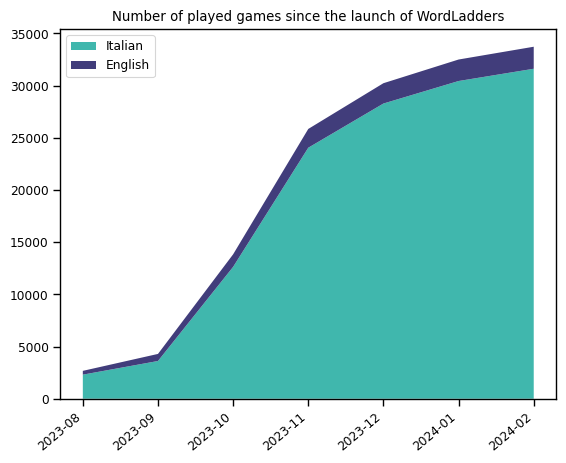}
  \caption{Played games per month.}
  \label{fig:sub4}
\end{subfigure}
\caption{Cumulative counts for number of users (a) and number of played games (b).}
\label{fig:growth}
\end{figure*}

In the first six months since its official launch at the end of August 2023, the application has seen continuous growth both in terms of users and number of games played, as we can see from the Figures \ref{fig:growth}. At the time of writing (13/02/2024), there have been over 30K games played from 3K active players\footnote{All data is collected in compliance with GDPR regulations, as described in the data policy (see \url{ANON_LINK}).}. Figure \ref{fig:pies} shows some demographic information about users (educational level and profession).

Table \ref{table:played} reports the rough counts of users and played games, showing that the most played game mode is \textit{single player}, which represents about 83\% of all the interactions with the app, while the other two game modes, i.e., \textit{challenge} and \textit{team}, gather together less than the 17\%. All plays produce a ladder of words from each player in every game mode.

Despite the instruction provided in the app, not all the submitted word ladders do reflect the taxonomic order the players are asked to produce (i.e., from the most generic on top to the most specific on bottom). Sometimes users tend to produce semantic relations other than hypernym/hyponym, as we can see from the \textit{bad ladders} reported in Table \ref{tab:word_ladders_examples}, which are examples of free association. While totally useless data are easy to filter out (e.g. non-words, blasphemy, mispellings, etc.) with simple rule-based approaches, it is more difficult to discriminate among word ladders that contain the desired hierarchical semantic relation among the words from those ladders that contain other semantic relations. Current ongoing work is aimed at disentangling the different semantic relations introduced by users within their played matches. The semantic annotations of word-to-word relations within each produced ladder will also be used to compare users' abilities to identify and discriminate between semantic relations.

\section{Conclusion}

The data collection through Word Ladders is currently ongoing, as previously illustrated. The final objective is to obtain around 100 played matches on each of the words given by the application. The present paper provided an overview of the application, its logic, the theoretical motivations that led to its implementation, the type of data collected, the data collection activities, and how the data will be utilized. Moreover, we illustrated the experimental approach tuned towards an educational objective of linguistic training, which we argue is a promising one for the motivation and involvement of students and pupils as well as general public in large-scale data collections in language sciences. As illustrated, Word Ladders has demonstrated its efficacy in raising users' motivation and engagement in data collection activities, in contrast to conventional experimental modalities such as online surveys. 

In the forthcoming months, our efforts will be directed towards expanding the dissemination of Word Ladders for the collection of English word ladders within educational contexts. This initiative aims to contribute to the development of parallel knowledge architectures in English and Italian languages, enriched with hierarchical semantic relations between words, supplemented by users' sociodemographic data. These resources will be used, as described, to measure word specificity and then investigate how categorical generalizations and specifications function in human cognition. Moreover, a parallel research aims at approximating the functioning of Word Ladders using LLMs, to explore to what extent categorical generalizations and specifications performed by LLMs compare to those performed in human cognition.

\section{Bibliographical References}\label{sec:reference}

\bibliographystyle{lrec-coling2024-natbib}
\bibliography{lrec-coling2024-example}

\end{document}